\documentclass{article}

\usepackage{microtype}
\usepackage{graphicx}
\usepackage{subfigure}
\usepackage{booktabs} % for professional tables

\usepackage{comment}
\usepackage{amsmath}
\usepackage{amssymb}
\usepackage{multirow}
\usepackage{tikz}
 
\usepackage[toc,page]{appendix}
\usepackage{placeins}
\usepackage[ruled,vlined]{algorithm2e}
\usepackage{xcolor}
\usepackage{bm}

\definecolor{grey}{RGB}{105,105,105}
\definecolor{ForestGreen}{RGB}{145, 209, 62}

\usepackage{hyperref}

\newcommand{\bu}{\mathbf{u}}

\newcommand{\bx}{\mathbf{x}}

\newcommand{\bbE}{\mathbb{E}}

\usepackage[accepted]{icml2021}

\icmltitlerunning{Measuring Fairness in Generative Models}

\begin{document}

\twocolumn[
\icmltitle{Measuring Fairness in Generative Models}



\icmlsetsymbol{equal}{*}

\begin{icmlauthorlist}
\icmlauthor{Christopher T.H Teo}{to}
\icmlauthor{Ngai-Man Cheung}{to}
%\icmlauthor{Aeiau Zzzz}{equal,to}
%\icmlauthor{Bauiu C.~Yyyy}{equal,to,goo}
\end{icmlauthorlist}

\icmlaffiliation{to}{Information System Technology and Design, Singapore University of Technology and Design, Singapore}
%\icmlaffiliation{to}{Department of Computation, University of Torontoland, Torontoland, Canada}
%\icmlaffiliation{goo}{Googol ShallowMind, New London, Michigan, USA}
%\icmlaffiliation{ed}{School of Computation, University of Edenborrow, Edenborrow, United Kingdom}

\icmlcorrespondingauthor{Christopher$\_$teo@mymail.sutd.edu.sg}
%\icmlcorrespondingauthor{Cieua Vvvvv}{c.vvvvv@googol.com}
%\icmlcorrespondingauthor{Eee Pppp}{ep@eden.co.uk}

\icmlkeywords{Fairness, bias mitigation, metric, GAN, ICML}

\vskip 0.3in
]



\printAffiliationsAndNotice{}  % leave blank if no need to mention equal contribution
\begin{abstract}
Deep generative models have made much progress in improving training stability and quality of generated data.
Recently there has been increased interest in the fairness of deep-generated data. 
Fairness is important in many applications, e.g. law enforcement, as biases will affect efficacy.
Central to fair data generation are the fairness metrics for the assessment and evaluation of different generative models.
In this paper, we first review fairness metrics proposed in previous works and highlight potential weaknesses. We then discuss a performance benchmark framework along with the assessment of alternative metrics.
\end{abstract}
\section{Introduction}
%General Introduction==================================================
Generative models have been well researched since the introduction of the Variational-Autoencoder (VAE) \cite{kingmaAutoEncodingVariationalBayes2014} and Generative Adversarial Network (GAN) \cite{NIPS2014_5ca3e9b1,goodfellowNIPS2016Tutorial2017}. Focusing on GANs, much research has been targeted at improving model architecture and performance, e.g. StyleGAN \cite{karrasStyleBasedGeneratorArchitecture2019} and BIGGAN \cite{brockLargeScaleGAN2019}, or finding solutions to stability issues, e.g. balancing discriminator-generator and mode collapse \cite{metzUnrolledGenerativeAdversarial2017,gulrajaniImprovedTrainingWasserstein2017, salimansImprovedTechniquesTraining2016, Tran2018DistGANAI,NEURIPS2019_d04cb95b}, improving data-efficiency \cite{9319516}, and detectability of deep-generated images \cite{Chandrasegaran_2021_CVPR}.
However, to our knowledge, little research has gone into addressing biases. This is an important factor to consider as it limits the potential of generative models. For instance, in a GAN application on the facial composition of criminal profiles \cite{9137812}, biases in gender may result in wrongful profiling.

{\bf In this work,}
we take a closer look at the fairness metrics 
for evaluating deep generative models.
A deep generative model $G_\theta$ produces synthetic data $\bx$. 
%The synthetic data 
$\bx$ follows some model distribution 
$q_{\theta}$ imposed by the generator network.
In many cases, $\bx$ 
can %could be 
biased w.r.t. some 
{\em targeted attribute}  $\bu \in \mathbb{R}^k$.
%Note that 
$\bu$ is a one-hot vector representation of the targeted attribute, and $k$ is the cardinality of the attribute, e.g. $k = 2$ if 
%the 
$\bu$ corresponds to gender, or $k = 4$ if $\bu$ is a compound attribute corresponding to gender and two different hair colours.

In many cases, $\bu$ is a  {\em latent} attribute.
Therefore, to evaluate the fairness of the generated data $\bx$  w.r.t. $\bu$, one would need to determine the attribute value through an  {\em attribute classifier}, $C$ \cite{groverBiasCorrectionLearned2019,choiFairGenerativeModeling2020,tanImprovingFairnessDeep2020}.
In particular, 
for an observed $\bx$,
the attribute classifier $C$ produces a soft output  
$\bar{p}_{\theta'}(\bu|\bx) = C(\bx)$.
Then, some discrepancy measure $D(.,.)$ between $\bbE_{x \sim q_{\theta}}[C(\bx)]$ and the uniform probability vector $\bar{p}$ can be used to quantify the fairness of $G_\theta$, where
 $\bar{p} = [\frac{1}{k}, \frac{1}{k}, \cdots ]$.
 %Notes: We explicitly address the definintion of fairness in the following sentence as per milan's comments that fairness is broad and this definition need not be "fair" when it comes to real world representation.
As per previous works \cite{choiFairGenerativeModeling2020,xuFairGANFairnessawareGenerative2018,tanImprovingFairnessDeep2020}, we utilise a uniform distribution to determine if the model has the ability to achieve statistical parity \cite{catonFairnessMachineLearning2020} - each equal probability is given to each outcome.%- each group receives an equal probability of outcome.
For example, in \cite{choiFairGenerativeModeling2020}, the 
L2 norm is used to measure the fairness discrepancy (FD) and is given by:
\begin{equation}
f = D(\bar{p},\bbE_{x \sim q_\theta} [C(\bx)]) 
= |\bar{p} - \bbE_{x \sim q_\theta} [C(\bx)] |_2
    \label{eqn:reweight10}
\end{equation}
If $f=0$, then $G_\theta$ is considered to be perfectly fair w.r.t. the targeted attribute $\bu$.
%Linking the relevance of poor C to f and the need to study it 
On the other hand, taking a second look at (\ref{eqn:reweight10}), one may realize that $f$ is highly dependent on $C$. Specifically, error in $C$ is inevitable in practice. However, its effect on $f$ has not been investigated. Even with a perfectly fair $G_\theta$, we may not obtain $f=0$ due to error in $C$.

\textbf{Related work:} A few other limited works have also tried to quantify fairness in GAN. \cite{xuFairGANFairnessawareGenerative2018} measures statistical parity with an additional discriminator. \cite{tanImprovingFairnessDeep2020} on the other hand, utilises a similar method as \cite{choiFairGenerativeModeling2020} with the addition of determining if the contextual attributes $u'$ of the generated images are maintained w.r.t to the population distribution. As a result of Xu's dependency on the model's architecture, we focus on the works of \cite{choiFairGenerativeModeling2020,tanImprovingFairnessDeep2020} whose use of an \textit{auxiliary classifier} $C$ is deemed advantageous due to its ease of deployment.

%Brief explanation of the study 
To understand how different choices of $D$ may affect the validity of $f$ under errors in $C$, we conduct an empirical study in this work. 
The main idea of our study is as follows. Given a deep generative model $G_{\theta}$ (biased or fair w.r.t. $\bu$) and an \textit{attribute classifier} $C$, we examine different 
choices of 
discrepancy measure $D$. We compute the value $f = D(\bar{p},\bbE_{q_\theta} [C(\bx)])$, and compare it with the {\em ground-truth} discrepancy value $f^* = D(\bar{p},\bbE_{q_\theta} [C^*(\bx)])$, where $C^*$ is an attribute classifier with perfect accuracy. The difference between $f^*$ and $f$ indicates the validity of selecting $D$  under attribute classification errors in $C$.

\textbf{Our contributions are:}
\begin{itemize}
    \item Identification of the effects of inaccuracies in $C$ on fairness metrics. 
    \item A methodological framework to analyse and quantify the differences in fairness metrics for generative models, and a recommendation of a robust metric.
\end{itemize}

%\section{Setup for Analysis and experiments} %Problem Setup for analysis
\section{Problem Setup and Analysis}
%\subsection{Classifier \& Metrics}
%Notes(Camera ready) Commenting out this section as it may cause confusion in contributed works/past: 
%We later introduce a selection of $D$ belonging to 3 different categories amorphic e.g. L1, L2 and Wasserstein (WS) distance, morphic e.g. specificity(sp) and hybrid e.g. information specificity(IS). Each category quantifies similarity differently, some more robust to the volatility introduced by $C$ inaccuracies. Further discussions in section 4.

%Introduction of different Sampling technique + obtaining f^* and f (Added 090621)
%Problem setup
%\subsection{Sampling Psuedo-generated Data }
\subsection{Experiment Setup}
In our experiments we utilise ResNet-18 \cite{heDeepResidualLearning2015} as our $C$, training it on different $\bu$ and $k$ configurations. We use Adam as our optimiser with $lr=10e^{-3}$. 
Suppose that for a dataset, the ground-truth distribution of some attributes is denoted by $\bar{p}_{\theta}^{*}$. In order to evaluate the effect of inaccuracies in \emph{C} on a given discrepancy measure \emph{D} in a controlled setting, first we need to simulate a generator $G_{{\theta}^\prime}$ that generates the data with assumed attribute distribution $\bar{p}_{\theta}^{*}$. To this end, we simply construct a dataset by sampling an available generated dataset  $D_{pp}$ (e.g. CelebA \cite{liuDeepLearningFace2015}). The sampling helps to not concern ourselves with the quality or diversity of $G_{{\theta}^\prime}$. For example, if $\mathbf{u}$ corresponds to gender ($k=2$), and $\bar{p}_{\theta}^{*}=[0.9, 0.1]$, given that $D_{pp}$ has 100 samples for each of male and female, we randomly sample 90 males and 10 females. Then we apply the attribute classifier and calculate the approximated distribution $\bar{p}_{{\theta}^\prime}$ which can be used to calculate $f=D(\bar{p},\bar{p}_{{\theta}^\prime})$.

Lastly, to mimic the perfect classifier $\bbE_{q_{\theta'}} [C^*(\bx)]$, we forgo the sampling and directly calculate $f^*$ via $\bar{p}^*_\theta$, i.e. $\bbE_{q_{\theta'}} [C^*(\bx)]=\bar{p}^*_\theta$, and hence $f^* = D(\bar{p},\bar{p}^*_{\theta})$. $f-f^*$ thus indicates the deviation of the fairness score as a result of inaccuracies in $C$.

%\section{Analysis of Previous Works}
\subsection{Analysis of Previous Works}
We emphasise a few key requirements that are necessary for effective FD measurement: 1) $\bu$ has to be well defined such that fair representation is achievable 2) $C$ has to have relatively high accuracy with low accuracy variability between the various $\bu$ 3) Appropriate metric has to be utilised such that the FD score is robust against noise such as inaccuracies in $C$ and selection of hyper-parameters, e.g. $k$. We focus on criteria 3 in this paper and highlight a few properties that need to be addressed.

%In the following, we clarify a few notations and terms.
We first clarify a few notations and terms.
%Concise version of the Fair-EP/AB-EP explanation
The extreme points (EP) in the distribution include the ideal worst case bias scenario which we call absolutely bias (AB-EP), and also the ideal fair scenario (Fair-EP). Using the previous gender example ($k=2$), there are two AB-EPs: $\bar{p}^*_{\theta-bias1}=[1,0]$, and $\bar{p}^*_{\theta-bias2}=[0,1]$, and the Fair-EP would be $\bar{p}^*_{\theta-fair}=\bar{p}=[0.5 ,0.5]$. Following this, the largest possible fairness discrepancy score achievable with $C^*$ is denoted by $f^*_{max}$, which is the $f^*$ for each of AB-EPs. Utilising our previous example, $f^*_{max} = D(\bar{p},\bar{p}^*_\theta)$, where $\bar{p}^*_\theta \in \{\bar{p}^*_{\theta-bias1},\bar{p}^*_{\theta-bias2}\}$. 
%Rephrased above-------------------------------------------------------------------
%Extreme points (EP) references the most extreme distribution, absolutely bias (AB-EP) and fair (Fair-EP) e.g.  $Attribute=\{Gender\}$, $k=2$, $\bar{p}=[\frac{1}{2},\frac{1}{2}]$,\; $\bar{p}^*_{\theta-bias1}=[1,0]$,\;$\bar{p}^*_{\theta-bias2}=[0,1]$,\;$\bar{p}^*_{\theta-fair}=[\frac{1}{2},\frac{1}{2}]$, AB-EP would refer to  $\bar{p}^*_{\theta-bias1}$ and $\bar{p}^*_{\theta-bias2}$ and fair-EP would refer to $\bar{p}^*_{\theta-fair}$. Next, $f^*_{max}$ is the largest possible fairness score achievable with $C^*$. This $f^*_{max}$ is the $f^*$ for AB-EP, the ideal worst case bias scenario e.g. $k=2$ , $f^*_{max} = D(\bar{p},\bar{p}^*_\theta)$, $\bar{p}^*_\theta$=\{[1,0],[0,1]\} .
%-----------------------------------------------------------------------------------
 Note that $f^*\ge f$.
With that, we highlight a few weaknesses of the past works with reference to FD (\ref{eqn:reweight10}) and propose solutions that would mitigate these weaknesses.\\ \\
%Upper Bound Scale Issue of metric (1)
\textbf{a) Scale:} The current metrics do not have a consistent upper bound, thereby making experimental comparison difficult.
%Which by definition of similarity measures \cite{charfiPossibilisticSimilarityMeasures2020} would not constitute it as a measure. 
For example, as attribute size increases, the $f^*_{max}$ increases, i.e. $k=2$, $f^*_{max}=0.707$ and $k=4$, $f^*_{max}=0.866$. Hence, we require a metric that has a fixed scale which does not vary with hyper-parameter changes. \\
\textbf{Solution:} As an easy fix, we propose to normalise each of our proposed metrics with their $f^*_{max}$, such that FD score $\in [0,1]$. As per the previous $k=2$ example, if given a $\bar{p}^*_{\theta}=[0.9,0.1]$ where $f^*=0.565$ and $f^*_{max}=0.707$, our normalised score is $f^*_{Norm}=\frac{0.565}{0.707}=0.799$. Note that each metric with different $k$ would have a different normalisation factor $N_{factor}$, as per Annex A.1 Table \ref{tab:norm}. This normalisation would fix all metrics' scales $\in [0,1]$, allowing comparison between them. We thus utilised this normalisation technique for all subsequent FD scores, where all $f$ and $f^*$ can be assumed to have been normalised.

\begin{figure}[ht]
\begin{center}
\centerline{\includegraphics[width=\columnwidth]{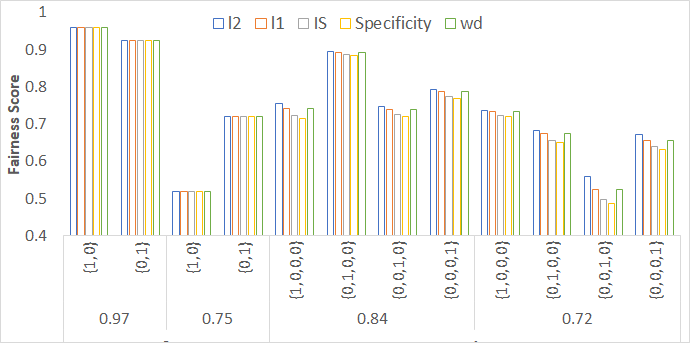}}
\caption{The effect of the accuracies of $C$ on different fairness metrics at AB-EP. X-axis sections: different $C$ as per (2.2b) Row1: AB-EPs with different $\bu$, Row 2: Accuracies.}
\label{fig:abepacc}
\end{center}
\end{figure}

\begin{figure}[ht]
\begin{center}
\centerline{\includegraphics[width=\columnwidth]{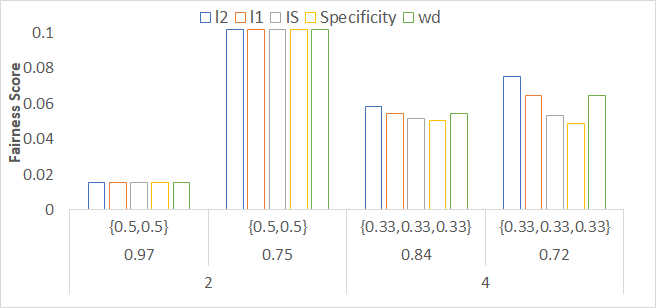}}
\caption{The effect of the accuracies of $C$ on fairness metrics at Fair-EP. X-axis sections: different $C$ as per (2.2b) Row1: AB-EPs with different $\bu$, Row 2: Accuracies.}
\label{fig:fairepacc}
\end{center}
\end{figure}

%Inaccuracy in $C$
\textbf{b) Inaccuracy in $C$:}
The imperfect $C$ presents a challenging problem where inaccuracies are propagated to the FD scores, making the measure unreliable. We demonstrate this by comparing the normalised FD score, of our proposed metrics introduced later in section 3, on four different $C$ of various accuracies and $k$ configurations. These $C$ were trained on the CelebA data set whose attributes include 1) Gender 2) Youth 3) Male and black-hair 4) Young and smiling.\\
%To demonstrate this, we utilise CelebA data set to train $C$ of various accuracies and $k$. These attributes include 1) Gender 2) Youth 3) Male and black-hair 4)Young and smiling. \\
%Pinching effect-----------------------------------------------------------------------------------------
\textbf{Pinching effect:} In Fig \ref{fig:abepacc} and \ref{fig:fairepacc} we observe that a decrease in the accuracy of $C$ results in a "pinching effect" whereby at fair-EP, $f$ increases and at AB-EP, $f$ decreases. This causes deviation from the theoretical optimal score of 0 and 1. This is in line with our intuition that a less accurate $C$ tends towards a random classification, resulting in $\bar{p}_{\theta'}$ having a more uniform distribution. \\
%Internal Variability------------------------------------------------------------------------------------
\textbf{Internal Variability:} Next, there exists internal variability where different $\bu$ have different classification accuracies, e.g. Fig.\ref{fig:abepacc} the first classifier on the left has accuracies of 0.98 and 0.95 for attributes [1,0] and [0,1] respectively. These varying accuracies form a bias that propagates to the FD score. Thus $\bar{p}_{\theta'}$ of the same shape but on different supports may produce different scores, e.g. in Fig \ref{fig:abepacc}, each AB-EP measures different scores even though $f^*=1$. This internal variability worsens as $k$ increases due to the increased difficulty to train $C$ (see Annex A). Hence, we require a metric that is robust to these inaccuracies and whose measurements are close to $f^*$. 
%Both L2 and KL divergence are not optimal metrics which we will discuss shortly. 

%As such, we emphasise a few key requirements that are necessary for effective FD measurement: 1) $\bu$ has to be well defined such that fair representation is achievable 2) $C$ has to have relatively high accuracy with low accuracy variability between the various $\bu$ 3) Appropriate metric has to be utilised such that the FD score is robust against noise such as inaccuracies in $C$ and selection of hyper-parameters, e.g. $k$. We focus on criteria 3 in this paper and proceed to the proposed metric to mitigate the problems mentioned in 2.2.

%------------------------------------------------------------------
\section{Proposed Metrics}
\label{section:metrics}
Generative models cannot utilise the traditional measurement of fairness as classifiers, e.g. Equalised Odds, Equalised Opportunity \cite{hardtEqualityOpportunitySupervised2016} and Demographic Parity \cite{feldmanCertifyingRemovingDisparate2015}, as a result of their different objectives.
Instead, we can evaluate the fairness metric as a problem of measuring similarities between probability distribution $\bar{p}$ and $\bar{p}_{\theta'}$, i.e. $D(\bar{p},\bar{p}_{\theta'})$. Similarities refer to how the shape of the distributions resemble one another. 
In this section, we explore the various $D$ utilised to describe this similarity. \cite{charfiPossibilisticSimilarityMeasures2020} differentiate these similarity measures into two categories, amorphic and morphic. 

%Notation clarification\
The following notations $\bar{p}_{\theta'_i}$ and $\bar{p}_{i}$ denotes the $i^{th}$ outcome in the distribution, e.g.  $k=4, \bu=\{[0,0],[0,1],[1,0],[1,1]\}$, $\bar{p}_{\theta'}=[0.3,0.1,0.2,0.4]$ then $\bar{p}_{\theta'_1}=0.3$ would be the probability for $\bu=[0,0]$.
%The following notations $\bm{\pi_1}$ and $\bm{\pi_2}$ denote probability vectors for $\bu$, where $\bm{\pi_1},\bm{\pi_2}\in \mathbb{R}^k$, e.g. gender and black hair, $k=4$ $\bu=\{[0,0],[0,1],[1,0],[1,1]\}$ $\bm{\pi_1}=[0.3,0.1,0.2,0.4]$, where the latter is the probability distribution of the former, i.e. $\pi_{1i}=0.3$ of $\bu=[0,0]$.
\emph{Amorphic metrics} are direct ``point-to-point" measurements between distributions, e.g. Normalised Manhattan Distance (L1), Normalised Euclidean Distance (L2) and Wasserstein Distance (WD)/Earth-mover Distance. 
%that solves an optimisation problem for the minimum work for one distribution to follow another.
%L1

% {\bf \color{red} Man }
% are these just $D(,)$ in the introduction? if so, pls connect them. also the eqn for L1 seems incorrect? are $\pi$  probability vectors for $\bu$ if so why sum over $n$?
%L1
\begin{equation}
%   L1(\bm{\pi_1},\bm{\pi_2})=\frac{1}{k}\sum^k_i|(\pi_{1i}-\pi_{2i})|
   %Amended for consistent notation
   L1(\bar{p}_{\theta'},\bar{p})=
   \frac{1}{k}\sum^k_i|(\bar{p}_{\theta'_i}-\bar{p_i})|
    \label{eqn:L1}
\end{equation}
%L2
\begin{equation}
    % \begin{aligned}
   % L2(\bm{\pi_1},\bm{\pi_2})=\frac{1}{k}\sqrt{\sum_{i=0}^k\delta_i(\bm{\pi_1},\bm{\pi_2)}},\ 
   % \delta_i(\bm{\pi_1},\bm{\pi_2})=[{\pi_{1i}}-{\pi_{2i}}]^2
   %Amended for consistency
       L2(\bar{p}_{\theta'},\bar{p})=\frac{1}{k}\sqrt{\sum_{i=0}^k\delta_i(\bar{p}_{\theta'},\bar{p})},\ 
    \delta_i(\bar{p}_{\theta'},\bar{p})=[\bar{p}_{\theta'_i}-{\bar{p}_{i}}]^2
    % \end{aligned}
    \label{eqn:L2}
\end{equation}
%Wasserstein distance
\begin{equation}
\begin{aligned}
    %WD(\bm{\pi_1},\bm{\pi_2})=\inf_{w\in \mathbb{R}^{nxm}} \sum_{i=0}^n %\sum_{j=0}^m w_{ij}d(\pi_{1i},\pi_{2j})\\
    %s.t\; \sum_{j=0}^m w_{i,j}=\pi_{1i} \forall i, \; \sum_{i=0}^nw_{i,j}=\pi_{2j} \forall j
    %Amended for consistency
        WD(\bar{p}_{\theta'},\bar{p})=
        \inf_{w\in \mathbb{R}^{nxm}} \sum_{i=0}^k \sum_{j=0}^k w_{ij}d(\bar{p}_{\theta'_i},\bar{p}_j)\\
    s.t\; \sum_{j=0}^k w_{i,j}=\bar{p}_{\theta'_i} \forall i, \; \sum_{i=0}^k w_{i,j}=\bar{p}_j
    \forall j
    \label{eqn:wd}
\end{aligned}
\end{equation}

On the other hand, \textit{morphic metrics} describes the shape of the graph through quantifying relevant information in the distributions. For instance, Specificity Measurements (\ref{eqn:sp}) \cite{charfiPossibilisticSimilarityMeasures2020},  describes the variability in the distribution. $\Delta Specificity$ (\ref{eqn:deltasp}) then describes the similarity measure, the difference between the distributions' variability. However, for simplicity, we refer to this metric as specificity, since $sp(\bar{p})=0$. 

%Specificity
\begin{equation}
\begin{aligned}
    %sp(\bm{\pi_1})=\pi_{11}-\sum_{j=2}^k\alpha_j\pi_{1j} \quad s.t \quad  
    %\alpha_j=\frac{n-j}{\sum_{j=2}^nj} \;, \; \\ 
    %\sum_{j=2}^n\alpha_j=1 \ , \ 
    %\alpha_j>\alpha_i\;, \;  \pi_{11} \ge \pi_j \ge \pi_i  \;, \; j<i 
    %Amended for consistency
    sp(\bar{p}_{\theta'})=\bar{p}_{\theta_1'}-\sum_{j=2}^k\alpha_j\bar{p}_{\theta_j'} \quad s.t \quad  
    \alpha_j=\frac{k-j}{\sum_{j=2}^kj} \;, \; \\ 
    \sum_{j=2}^k\alpha_j=1 \ , \ 
    \alpha_j>\alpha_i\;, \;  \bar{p}_{\theta_1'} \ge \bar{p}_{\theta_j'} \ge \bar{p}_{\theta_i'}  \;, \; j<i   
\end{aligned}
\label{eqn:sp}
\end{equation}
\begin{equation}
    \Delta Specificty=|sp(\bar{p}_{\theta'})- sp(\bar{p})|
    \label{eqn:deltasp}
\end{equation}

A hybrid measure also exists, that take the combination of morphic and amorphic metrics, e.g. information specificity (IS) (\ref{eqn:is})\cite{charfiPossibilisticSimilarityMeasures2020} utilises a combination of the L1's amorphic point-to-point measurement to determine the displacement between the two distribution as well as specificity that measure the difference in distributions' variation. We set $\alpha=0.5$ for (\ref{eqn:is}), in our experiments.

%info-specificity
\begin{equation}
\begin{aligned}
    %information\_Specificity(\bm{\pi_1},\bm{\pi_2})=\\
    %{\alpha* L1(\bm{\pi_1},\bm{\pi_2})+(1-\alpha)*|sp(\bm{\pi_1})- sp(\bm{\pi_2})|}
    %Amended for consistency
    information\_Specificity(\bar{p}_{\theta'},\bar{p})=\\
    {\alpha* L1(\bar{p}_{\theta'},\bar{p})+(1-\alpha)*|sp(\bar{p}_{\theta'})- sp(\bar{p})|}
\end{aligned}
\label{eqn:is}
\end{equation}
 We did not consider KL-divergence as per \cite{tanImprovingFairnessDeep2020} as it resulted in computational problems when the support of $\bar{p}$ and $\bar{p}_{\theta'}$ were different, e.g. at AB-EP. In addition, it's non-symmetrical properties poses other problem beyond the scope of this paper. 

%Performance Benchmarks------------------------------------------------------
\subsection{Performance Benchmark} 
We utilise the following benchmarks to identify the ideal metric that is robust against inaccuracies in $C$ and hyper-parameter changes.

\textbf{Mean extreme point error (MEPE)} (\ref{eqn:epefair}) and (\ref{eqn:mepeab}) measures the metrics' deviations from the theoretical boundaries, 0 and 1 respectively. Note that $MEPE_{AB}$ is an average across multiple AB-EP. $s=\{\bar{p}_{\theta_{(1)}'},..,\bar{p}_{\theta_{(N)}'}\}$ denotes a set of approximated AB-EP/fair-EP distribution classified by $C$ where $N$ is the number of fair-EP/AB-EP in the set. For example k=2 and 4, when calculating $MEPE_{AB}$ we utilise $\bu$=\{[1,0],[0,1]\} and $\bu$=\{[1,0,0,0],[0,1,0,0],[0,0,1,0],[0,0,0,1]\} hence, $N=6$. We mix the errors across different $k$ in order to find a metric independent of $k$.
%Mean extreme point error (Fair)
\begin{equation}
    %MEPE_{fair}(D_{gen})= E_{fair}|f[P(D_{gen})]-0|
    MEPE_{fair}(s)= \frac{1}{N}\sum_{i=1}^N|D(\bar{p},\bar{p}_{\theta_{(i)}'};C)]-0|
    \label{eqn:epefair}
\end{equation}
%Mean extreme point error (bias)
\begin{equation}
    %MEPE_{AB}(D_{gen})= E_{AB}|f[P(D_{gen})]-1|] \\
     MEPE_{AB}(s)= \frac{1}{N}\sum_{i=1}^N|D(\bar{p},\bar{p}_{\theta_{(i)}'};C)]-1|
    \label{eqn:mepeab}
\end{equation}
\textbf{Extreme point variability ($EP-var_{(fair/AB}$)} (\ref{eqn:epv}) helps indicate the overall stability of the metrics by measuring the variability of $f$ at fair-EP/AB-EP. $\mu$ indicates the average FD score.
%Extreme point Variability
\begin{equation}
\begin{aligned}
   EP-var_{(fair/AB)}= \frac{\sum_{1}^{N}(D(\bar{p},\bar{p}_{\theta_{(i)}'};C)-\mu)^2}{N}\\ %,\,Ats\in\{2,4,8,16\} 
 %  ,\,N=Number\_of\_EP\\
\end{aligned}
\label{eqn:epv}
\end{equation}
\textbf{Mean error measurement (MEM)} (\ref{eqn:mem}) determines how far, on average, the approximated score utilising $C$ deviates from the theoretical ideal score. $s=\{\bar{p}_{\theta_{(1)}'},..,\bar{p}_{\theta_{(N)}'}\}$ and $s^*=\{\bar{p}^*_{\theta_{(1)}},..,\bar{p}^*_{\theta_{(N)}}\}$ denotes the set of approximated and ground-truth sample distributions respectively.
%as we sweep the distribution per algorithm 1. The notation $s$ represents a set of distribution. 
This metric thus indicates the overall effects that external factors, e.g. classifier's accuracy has on the $D$.
%Mean error measurements
\begin{equation}
\begin{aligned}
   %MEM(x)= E_{x}|f[P(D_{gen\_i})]-f[Id(D_{gen\_i})]| \\
  % \{D_{gen\_1},..,D_{gen\_N}\}\in x
       MEM(s,s^*)= \frac{1}{N}\sum_{i=1}^N|D(\bar{p},\bar{p}_{\theta_{(i)}'};C)-D(\bar{p},\bar{p}^*_{\theta_{(i)}})| \\
   %\{P_{gen 1},..,P_{gen N}\}\in x
\end{aligned}
\label{eqn:mem}
\end{equation}
Note that (\ref{eqn:epefair}),(\ref{eqn:mepeab}),(\ref{eqn:epv}) are calculate across $k$=\{2,4,8,16\} in our experiments.
%Include** Equations with regards to these benchmarks

\section{Experiments}

%Note: Remove this section, felt that this section was 1) Non essential 2) Was not convincing that it was beneficial 
%%%%%%%%%%%%%%%%%%%%%%%%%%%%%%%%%%%%%%%%%%%%%%%%%%%%%%%%%%%%%%%%%%%%%%%%%%%%%
% \subsection{Non-experimental Analysis }
% \begin{figure}[ht]
% \begin{center}
% \centerline{\includegraphics[width=\columnwidth]{Figures/support variability effect/SVE4.png}}
% \caption{$k=4$, Support variability effects on different metrics with different supports}
% \label{fig:sve}
% \end{center}
% \end{figure}
% \textbf{Support Variability Effect(SVE)}: We first evaluated the theoretical effect of changing supports size, essential in identifying non-representation in $\bu$. In Fig\ref{fig:sve}, we increase the support for 3 sample points L1 remained generally constant for $k=4$ but decreases when $k=8$, more in annex A. Whereas, specificity and IS scores increases, which are non-ideal behaviour as more support indicate fairer representation. Hence, l2 followed by l1 are more ideal metrics.
%%%%%%%%%%%%%%%%%%%%%%%%%%%%%%%%%%%%%%%%%%%%%%%%%%%%%%%%%%%%%%%%%%%%%%%%%%%%%

%Metric table comparsion%%%%%%%%%%%%%%%%%%%%%%%%%%%%%%%%%%%%%%%%%%%%%%%%%%%%%%%%%%%%%%%%%%%%%%

\subsection{Experiment Setup}
Next, we conduct the experimental evaluation of the various metrics.
We again utilise CelebA to train the next set of $C$ for the remaining experiments. $Attributes=$\{gender, black hair, smiling and bangs\} are incrementally trained into 4 different $C$ of increasing $k$. Without loss of generality, the attributes are binary and scale $k$ exponentially, $2^{attribute\_count}$ after permutations.
In 4.2 and 4.3, we measure fairness metrics response to varying k=\{2,4,8,16\}. We first analyse the fairness metric on their EPs followed by varying the distribution from AB-EP to fair-EP, generated with algorithm 1. \\

\begin{figure*}[ht!]
\begin{center}
\centerline{\includegraphics[scale=0.23]{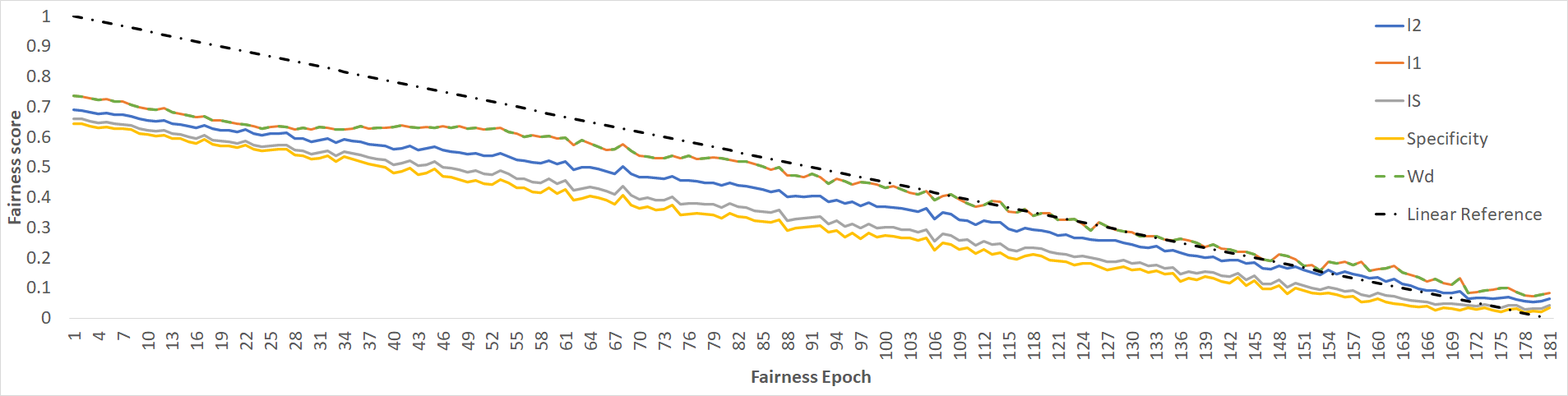}}
\caption{$k$=8, Normalised Fairness score against a sweeping distribution from AB-EP to Fair-EP}
\label{fig:8sweepscore}
\end{center}
\end{figure*}

\begin{figure*}[ht!]
\begin{center}
\centerline{\includegraphics[scale=0.19]{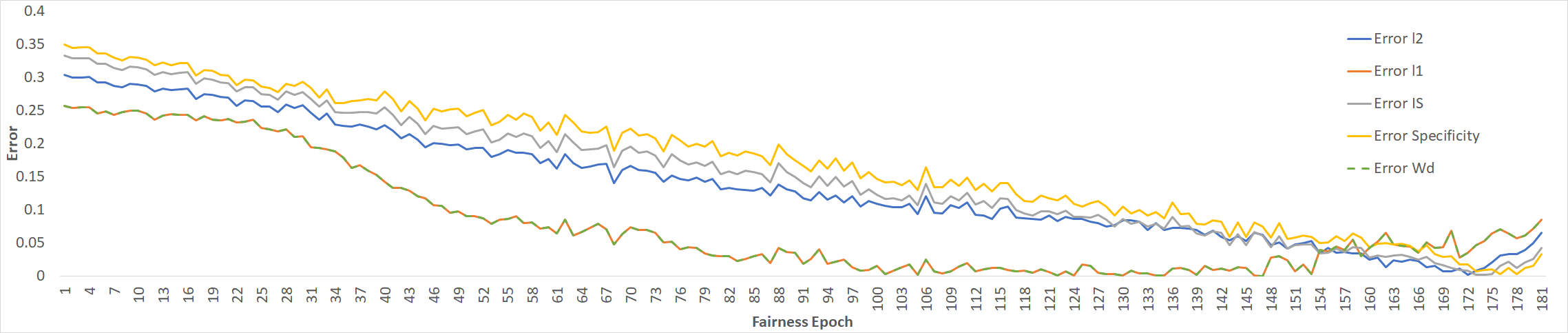}}
\caption{$k$=8 , Normalised Error score between the ideal score and the measured score ,$|f^*-f|$, against a sweeping distribution from AB-EP to Fair-EP}
\label{fig:8sweeperror}
\end{center}
\end{figure*}

\begin{algorithm}[H]
\SetAlgoLined
\KwResult {$arrayList$} \; 
 current\_dist=[100,0,0...0]\;\\
 uniform\_dist=$U(num\_of\_attr)$\; \\
 index=1\; \\
 arrayList\; \\
 \While{current\_dist != uniform\_dist}{
  \eIf{current\_dist[index]!=uniform\_dist[index]}{
   current\_dist[0]-step\\
   current\_dist[index]+step\\
   arrayList.append(current\_dist/100)
   }{
   index+1\;
  }
  
 }
 \caption{Distribution sweep from AB-EP to Fair-EP }
\end{algorithm}

\begin{table}[ht!]
    \centering
    \caption{Summary of the metric performances according to our performance benchmark. A lower score is better for all benchmarks. \textcolor{ForestGreen}{Green} indicates the best results and \textcolor{red}{Red} indicates worst }
    
    \resizebox{\columnwidth}{!}{\begin{tabular}{c|c|c|c|c|c}
    Benchmark Point&l2&L1&IS&Specificity&WD \\
    \hline \hline
    %Extreme point error---------------------------------------------------------
    \multicolumn{6}{c}{Mean Extreme point Error}\\
    \hline
    (Fair-EP) & 0.0588  &  \textcolor{red}{0.0851}  &  0.0361  &  \textcolor{ForestGreen}{0.0276}  &  \textcolor{red}{0.0851} \\
    (AB-EP) & 0.2106  &  \textcolor{ForestGreen}{0.1884}  &  0.2273  &  \textcolor{red}{0.2338}  &  \textcolor{ForestGreen}{0.1884} \\
    \hline
    %Extreme point Variability----------------------------------------------------
    \multicolumn{6}{c}{Extreme Point Variability}\\
    \hline
    (Fair-EP) & 0.0349  &  \textcolor{red}{0.0657}  &  0.0207  &  \textcolor{ForestGreen}{0.0168}  &  \textcolor{red}{0.0657} \\
    (AB-EP) & 0.1058  &  \textcolor{ForestGreen}{0.0698}  &  0.1149  &  \textcolor{red}{0.1212}  &  \textcolor{ForestGreen}{0.0698} \\
    \hline
    %Mean Extreme Error----------------------------------------------------
    \multicolumn{6}{c}{Mean Error Measurement}\\
    \hline 
    (2ATTR Sweep) & 0.0184 & 0.0184 & 0.0184 & 0.0184 & 0.0184\\
    (4ATTR Sweep) & 0.0884  &  \textcolor{ForestGreen}{0.0623}  &  0.0919  &  \textcolor{red}{0.1034} &  \textcolor{ForestGreen}{0.0623} \\
    (8ATTR Sweep) & 0.1386  &  \textcolor{ForestGreen}{0.0798}  &  0.1535  &  \textcolor{red}{0.1727}  &  \textcolor{ForestGreen}{0.0798} \\
     (16ATTR Sweep) & 0.1761  &  \textcolor{ForestGreen}{0.1155}  &  0.19  &  \textcolor{red}{0.2031}  &  \textcolor{ForestGreen}{0.1155}

    \end{tabular}}
    \label{tab:benchmarksummary}
\end{table}

\subsection{Extreme Points Analysis}
\textbf{Fair-EP vs AB-EP:} Overall, we observe as per Table \ref{tab:benchmarksummary} that fair-EP scores are closer to the theoretical value, i.e. lower error, than the AB-EP. This occurs due to \textit{trend-line effects}, which we will discuss in 4.3, as well as inter-variability discussed in 2.2, where a particular $\bu$ may have poorer accuracy and hence larger error in its AB-EP score, e.g. Fig \ref{fig:abepacc}, $\bu= [0,0,1,0]$. 
However, when measuring fair-EP any poor estimation by $C$ is averaged across all permutations of $\bu$ due to the uniform distribution being sampled. This contributes to the lower error measurement in $MEPE_{fair}$. Hence, this results in better estimation of the fair-EP with lower $EP-var_{fair}$ in comparison to $EP-var_{AB}$ as per Table 
\ref{tab:benchmarksummary}. We call this the \textit{internal variability effect} (see Annex D).\\

\begin{figure}[ht]
\begin{center}
\centerline{\includegraphics[width=\columnwidth]{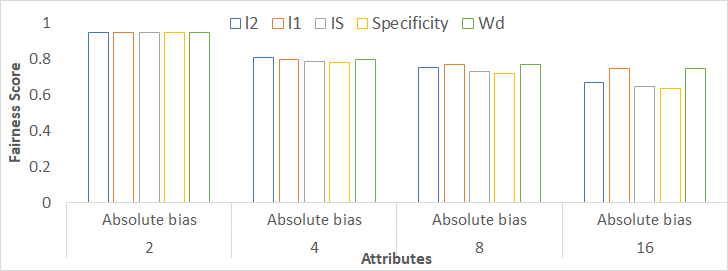}}
\caption{Mean normalised score for different $k$ at AB-EP. Theoretical optimum score=1, hence a larger score is better}
\label{fig:abepscore}
\end{center}
\end{figure}

\begin{figure}[ht]
\begin{center}
\centerline{\includegraphics[width=\columnwidth]{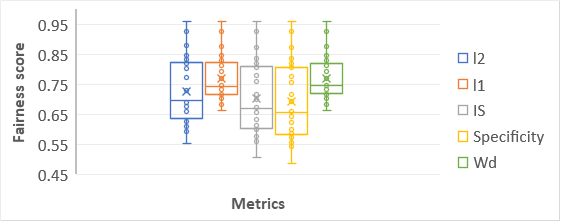}}
\caption{Overall variability of normalised scores at AB-EP for $k$= [2,4,8,16]}
\label{fig:abepvar}
\end{center}
\end{figure}
\textbf{AB-EP Analysis:} We utilise quantitative analysis to study the metrics behaviour of FD score and its variability at AB-EP, as seen in Fig \ref{fig:abepscore} and \ref{fig:abepvar} respectively. Specificity performs the worst with the highest $MEPE_{AB}$, regardless of $k$ size. It also has the highest $EP-Var_{AB}$, making it the least consistent fairness metric at AB-EP. Conversely, WD and L1 performed the best with the lowest $MEPE_{AB}$ and $EP-Var_{AB}$. As $k$ used is relatively small WD and L1 are generally identical to one another when the scores are rounded to 4 D.P.

We identify the advantages of the L1 metric in comparison to L2, specificity and by extension IS. 
%L1 explanation vs L2
L1's lower variability is attributed to the simplicity of the metrics linear scale, making it less susceptible to noise from volatility in $C$. Whereas, specificity on the other hand measures variability in $\bar{p}_{\theta'}$ and therefore is more sensitive to the noise prevalent at AB-EP. This influences the larger variance in its score as well as score deviations from $optimal\_max=1$. This volatility is further amplified by the decrease in accuracy as k increases. To validate this, Annex A Fig \ref{fig:abattrerror} shows specificity having the greatest increase in error when k increased. 

\begin{figure}[ht]
\begin{center}
\centerline{\includegraphics[width=\columnwidth]{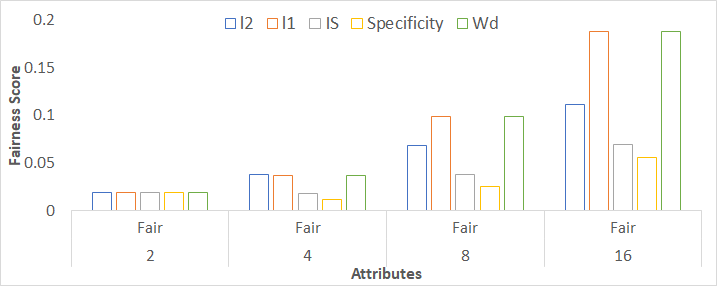}}
\caption{Mean normalised score for different $k$ at Fair-EP. Theoretical optimum score=0, hence a smaller score is better}
\label{fig:fairepscore}
\end{center}
\end{figure}

\begin{figure}[ht]
\begin{center}
\centerline{\includegraphics[width=\columnwidth]{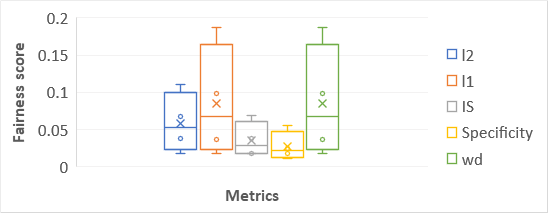}}
\caption{Overall variability of normalised scores at Fair-EP for $k$= [2,4,8,16]}
\label{fig:fairepvar}
\end{center}
\end{figure}

\textbf{Fair-EP Analysis:} Utilising Fig \ref{fig:fairepscore} and \ref{fig:fairepvar}, the metric performance at Fair-EP are the converse of AB-EP. L1/WD performs the worst and specificity the best.  
Amorphic metrics' have smaller $N_{factor} < 1$. As such their fairness scores are largely scaled up during normalisation, resulting in L1, L2 and WD relatively larger fairness scores and hence larger $MEPE_{fair}$. This scaling effect is amplified with increasing $k$, Annex A.1 Table \ref{tab:norm} for all the $N_{factors}$. Specificity on the other hand $N_{factor}=1$ regardless of the size of $k$, hence no scaling occurs, thereby attaining the lowest $MEPE_{fair}$. Furthermore, as per the internal variability effect, average accuracy is generally higher at the fair-EP, making specificity more stable 
%and hence contributing to its stability. Hence, specificity 
thereby attaining the lowest $EP-Var$. 

%Trend Analysis---------------------------------------------------------
\subsection{Metric Trend line Analysis}
The trend line analysis studies the behaviour of the metrics with increasing fairness distribution. Each increment in the fairness epoch tends towards a more uniform distribution. Fig \ref{fig:8sweepscore} shows that the metrics' general trends follow our expectation, where the highest score is seen at the beginning which decreases with each epoch. Overall L1/WD performed the best with the lowest MEM score and specificity the worst. 
A few interesting observations were seen in Fig.\ref{fig:8sweeperror} where 1)All metrics began with a large error that decreases with each fairness epoch and 2) L1's momentary increment in fairness score at epoch 38 to 52.
%and error particularly follows a steep decline. \\
%Addressing abnormalities in the trend (To add on more)

\textbf{General reduction in error:} The initial large error seen in Fig \ref{fig:8sweeperror} epoch 1, is the result of inaccuracy in $C$. 
This is a similar observation as discussed in 4.2 AP-EP analysis. Additionally, we observed a decrease in trend-line gradient, i.e. gradient $\propto$ $C$ accuracy which
%Thereby a decrease in AB-EP scores, as discussed in 5.2, and additionally a decrease in the trend-lines gradient. This 
creates deviation from its ideal trend, seen in Annex C Fig \ref{fig:idealsweep8}. 
The change in gradient then results in the gradual convergence towards the ideal scores, with each increment in fairness epoch, i.e. reduction in error. This same trend holds for $k$=4 and 16, seen in Annex C, where the former experiences less deviation as a result of its more accurate $C$ and the converse for the latter.
Furthermore, we note that when starting at different AB-EP the convergence rate are different, where the AB-EP whose $\bu$ has the lowest accuracy
%Furthermore, we note that each AB-EP converges at different rates where the point with the lowest accuracy 
converges the slowest (see Annex D). Lastly, we notice that not all metrics converge at the same rate, we address this in the following.

\textbf{L1 trend abnormality:} The increment in L1's fairness score is the result of the metric's ability to aggressively correct its error. We attribute this to L1's simplistic linear equation that is least susceptible to noise caused by inaccuracies in $C$ and hence the quickest to correct. This inference is supported by the trend that specificity, the metric most sensitive to poor accuracy in $C$, having the largest error and slowest convergence. Followed by IS who is influenced by specificity and finally L2, whose quadratic calculation creates larger variations from the true measurement as seen previously in 4.2.
The L1  aggressive correction makes it the most robust metric to poor accuracy in $C$. 
These trends are similarly observed in $k=4$ and $8$ as per Annex C. %where $k=4$ having higher accuracy observed steeper gradients, thereby following closer to the ideal trend line and the converse for $k=16$.
\section{Future works, Conclusion and Metric Recommendation}
    %Future works:
    \textbf{Future works: }our work defines fairness as a uniform distribution with the motivation of data augmentation where even representation of, e.g. hair colour, is ideal for balance learning . However, this definition may vary where fairness could mean "to follow the population distribution", e.g. the racial demographic of a country, which may not be uniform. Thus future work could explore this domain.
    
    %Conclusion
    \textbf{To conclude}, we have presented the existing problems in the current FD score and introduced a variety of different morphic, amorphic and hybrid fairness metrics to help mitigate the problem. We introduced performance benchmarks to determine the ideal metric. Through experimentation, we observed, that L1/WD metric performed the best at AB-EP and specificity at Fair-EP as per Table \ref{tab:benchmarksummary}. Furthermore, L1/WD demonstrates to be the most robust, having the lowest MEM. As such, there is no clear distinct optimal metric that favours all components of our benchmark. Thus, we recommend , hybrids metrics, IS to be used as the ideal middle ground. We also point out the unlikely-hood that the generated distribution will tend towards AB-EP. Hence, emphasis should be placed on Fair-EP, which IS does.
    
\section*{Acknowledgements}
We would like to thank Milad Abdollahzadeh for his constructive comments on the manuscript.

\bibliography{Main}
\bibliographystyle{icml2021}

%Appendix----------------------------------------------------------------------------------------------------
\newpage 
\onecolumn
\FloatBarrier
\begin{appendices}
Code: \url{https://github.com/Bearwithchris/Fairness_Metric}

%A------------------------------------------------------
% \section{Support variability}
% As $k$ increases, we observed that L1/WD experience a sharp change in fairness score when the fist new support was added, the first blue bar to the subsequent adjacent orange bar. This is as a result of the normalisation effect where the larger the $k$ , the bigger the normalisation effect.
% \begin{figure}[!ht]
% \centering
%     \includegraphics[scale=0.3]{Figures/support variability effect/SVE8.png}
%     \caption{Support variability for $k=16$}
%     \label{Study_Tree}
% \end{figure}

\FloatBarrier
%B------------------------------------------------------
\FloatBarrier
\section{Metric Study}
The following graph shows the variability of the $C$ accuracy between AB-EP as the dimension of $\bu$ increases. As previously discussed, the variability begins to increase in addition to the reduction in accuracy as the $k$ increases. This indicates that $C$ would have an \textbf{internal bias} to certain attributes.
\begin{figure}[!htb]
\centering
    \includegraphics[scale=0.4]{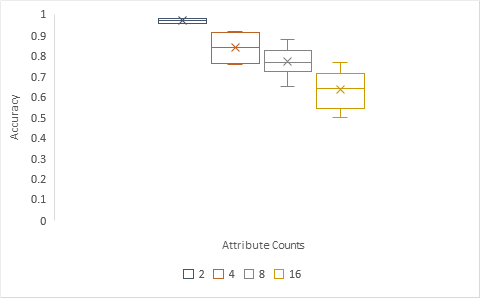}
    \caption{Accuracy Variability among different $C$ on different attributes }
    \label{Study_Tree}
\end{figure}

\begin{figure}[!htb]
\centering
    \includegraphics[scale=0.4]{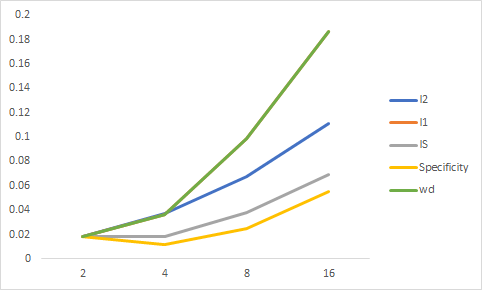}
    \caption{Extreme points Fair Error }
    \label{fig:fairattrerror}
\end{figure}

\begin{figure}[!htb]
\centering
    \includegraphics[scale=0.4]{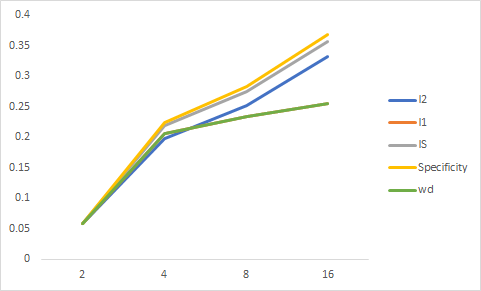}
    \caption{Extreme points AB Error }
    \label{fig:abattrerror}
\end{figure}

With this internal bias, we further in Fig \ref{fig:fairattrerror}, \ref{fig:abattrerror} we observe that the extreme point errors increase with $k$. Additionally, we begin to see a clear distinction between the metrics. In Fair-EP, specificity has the least error regardless of $k$, whereas WD begins diverging with increase error. However, the converse is seen for AB-EB. 

\FloatBarrier
\subsection{Normalisation factor}
The Normalisation factors are in Table \ref{tab:norm}, where the values in each cell represented the theoretical ceiling of each metric. Hence, the normalisation factor is an important addition to ensure that the fairness scores $\in[0,1]$ thereby allowing the metrics to have a direct comparison to one another. Furthermore, the metric would have little meaning as the theoretical ceiling changes according to attribute size. This implies that the fairness score could be synthetically improved simply by increasing $k$.  \\ \\
None-the-less we are aware that the the normalisation factor does create some problems. For example, in $k=8$ fair-EP we observe the raw metric to have the following scores $l2=0.00793$, $l1=0.02153 $, $IS=0.02316 $, $Specificity=0.02479 $ and $wd=0.02154 $. It is clear the L1 achieves a lower score that specificity in the raw score. However, as a result of the large difference in the normalisation factor upon normalisation $l2=0.06788 \;,\; l1=0.0984 \;,\; IS=0.03801 \;,\; Specificity=0.02479\;,\; wd=0.09848\;, \;wds=0.03801 $. Now, L1 is greater than specificity as a result of its smaller normalisation factor. \\ \\
On the other hand, when normalisation factor difference is small such as L2 and l1 we observe less significant effects. For example $k=16$ where the raw fairness scores are $l2=0.04093\; l1=0.09034\; IS=0.3664\; Specificity=0.6425\; wd=0.09034$ and Normalised scores $l2=0.6764\; l1=0.7709 \; IS=0.6559 \; Specificity=0.6425 \; wd=0.7709$. L1 remains larger than L2 regardless of L2's smaller normalisation factor.
\begin{table}[h!]
    \centering
    \caption{Normalisation factor}
    \begin{tabular}{c|c|c|c|c|c}
    Attributes  &  l2  &  l1  &  IS  &  Specificity  &  wd \\
    \hline
    2  &  0.353553391  &  0.5  &  0.75  &  1  &  0.5 \\
    4  &  0.216506351  &  0.375  &  0.6875  &  1  &  0.375 \\
    8  &  0.116926793  &  0.21875  &  0.609375  &  1  &  0.21875 \\
    16  &  0.060515365  &  0.1171875  &  0.55859375  &  1  &  0.1171875 \\

    \end{tabular}
    
    \label{tab:norm}
\end{table}
%C------------------------------------------------------
\FloatBarrier
\section{Classifier's attributes}
\begin{table}[h!]
    \centering
    \caption{Set 1 Classifiers for accuracy analysis}
    \begin{tabular}{c|c|c}
        Attributes & dimension of {$\bu$} & Accuracy  \\
        \hline
         Gender&2&0.98\\
         Youth&2&0.81\\
         Male,black hair&4&0.83\\
         young, Smiling&4&0.72\\
    \end{tabular}
    
    \label{tab:my_label}
\end{table}

\begin{table}[h!]
    \centering
    \caption{Set 2 Classifiers for attribute increment analysis}
    \begin{tabular}{c|c|c}
        Attributes & dimension of {$\bu$} & Accuracy  \\
        \hline
         Gender&2&0.98\\
         Gender, black-hair & 4 &0.86\\
         Gender,black hair, Smiling & 8 &0.78\\
          Gender, black hair, Smiling, bangs & 16 &0.66\\
    \end{tabular}
    
    \label{tab:my_label}
\end{table}
%D------------------------------------------------------

\FloatBarrier
\section{Trendline Sweep}

% \begin{algorithm}[H]
% \SetAlgoLined
% \KwResult {$D_{generate_list}$} \; 
%  current\_dist=[100,0,0...0]\;\\
%  uniform\_dist=$U(num\_of\_attr)$\; \\
%  index=1\; \\
%  arrayList\; \\
%  \While{current\_dist != uniform\_dist}{
%   \eIf{current\_dist[index]!uniform\_dist[index]}{
%   current\_dist[0]-step\\
%   current\_dist[index]+step\\
%   arrayList.append(current\_dist)
%   }{
%   index+1\;
%   }
  
%  }
%  \caption{Fairness distribution allocation}
% \end{algorithm}

\begin{figure*}[!ht]
\centering
    \includegraphics[scale=0.38]{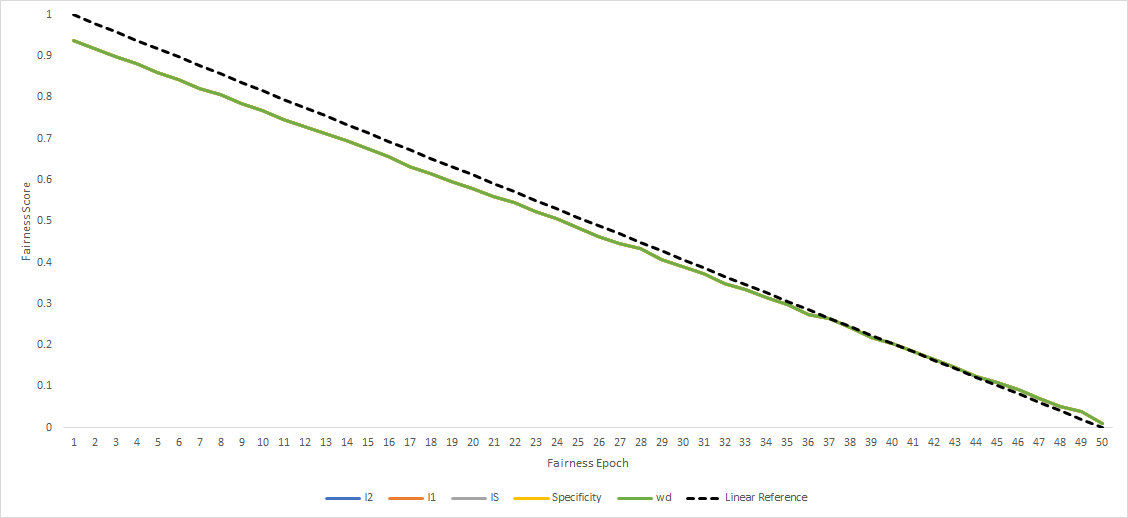}
    \caption{Normalise Fairness score with sweeping distribution from AB-EP to Fair-EP at k=2}
    \label{Study_Tree}
\end{figure*}

\begin{figure*}[!ht]
\centering
    \includegraphics[scale=0.4]{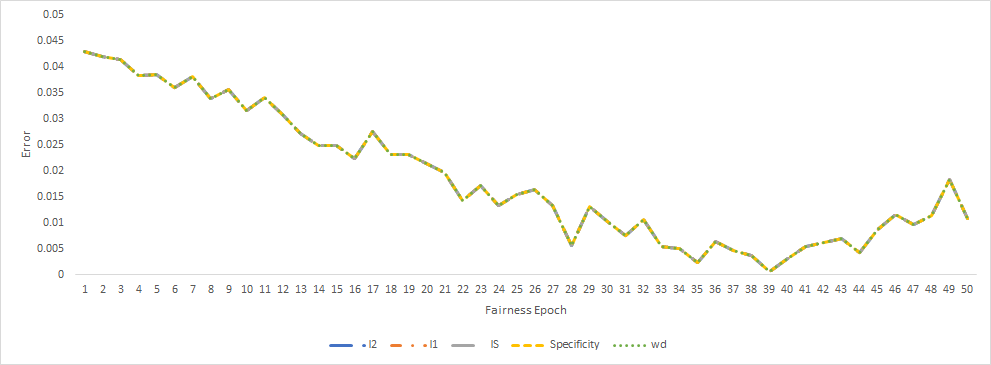}
    \caption{Error with sweeping distribution from AB-EP to Fair-EP at k=2}
    \label{Study_Tree}
\end{figure*}

\begin{figure*}[!ht]
\centering
    \includegraphics[scale=0.28]{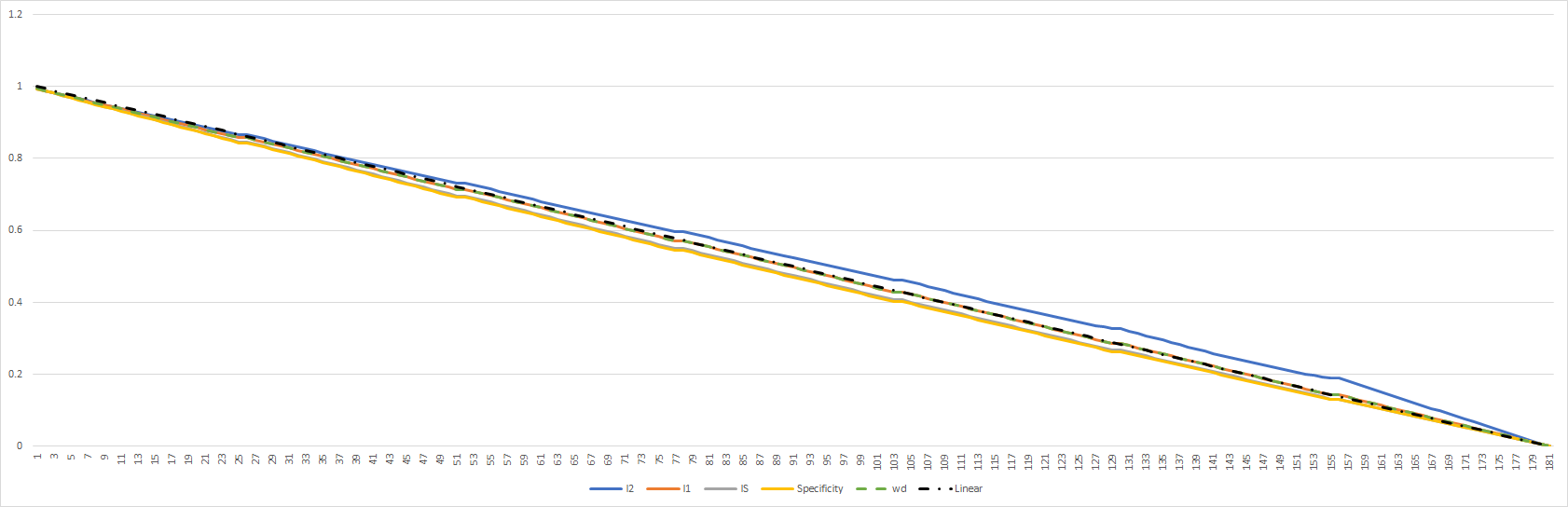}
    \caption{Ideal Normalise Fairness score with sweeping distribution from AB-EP to Fair-EP at k=4}
    \label{fig:idealsweep8}
\end{figure*}

\begin{figure*}[!ht]
\centering
    \includegraphics[scale=0.3]{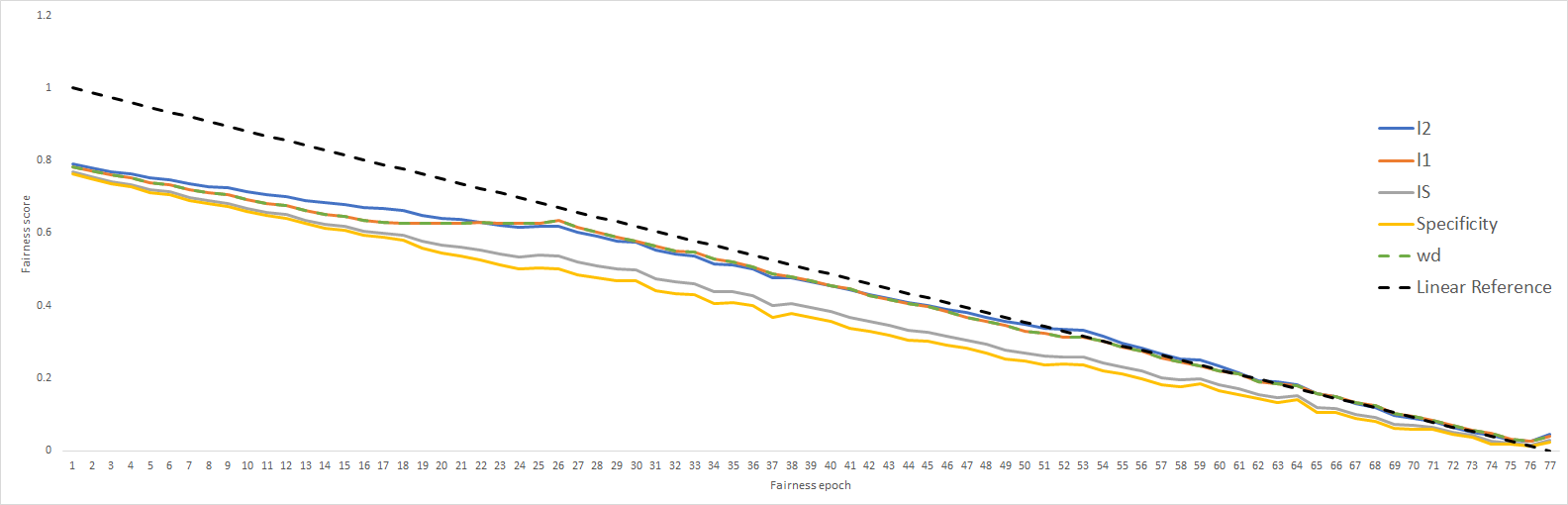}
    \caption{Normalise Fairness score with sweeping distribution from AB-EP to Fair-EP at k=4}
    \label{Study_Tree}
\end{figure*}

\begin{figure*}[!ht]
\centering
    \includegraphics[scale=0.3]{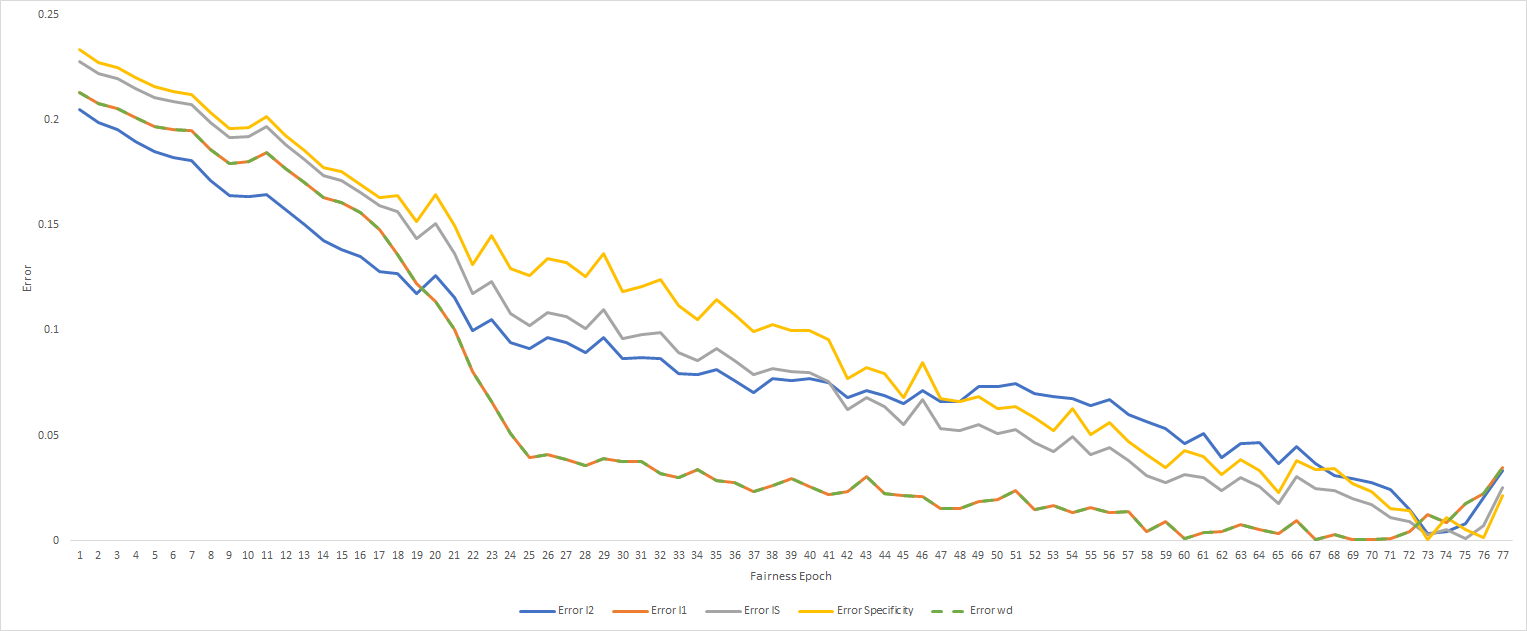}
    \caption{Error with sweeping distribution from AB-EP to Fair-EP at k=4}
    \label{Study_Tree}
\end{figure*}

\begin{figure*}[!ht]
\centering
    \includegraphics[scale=0.2]{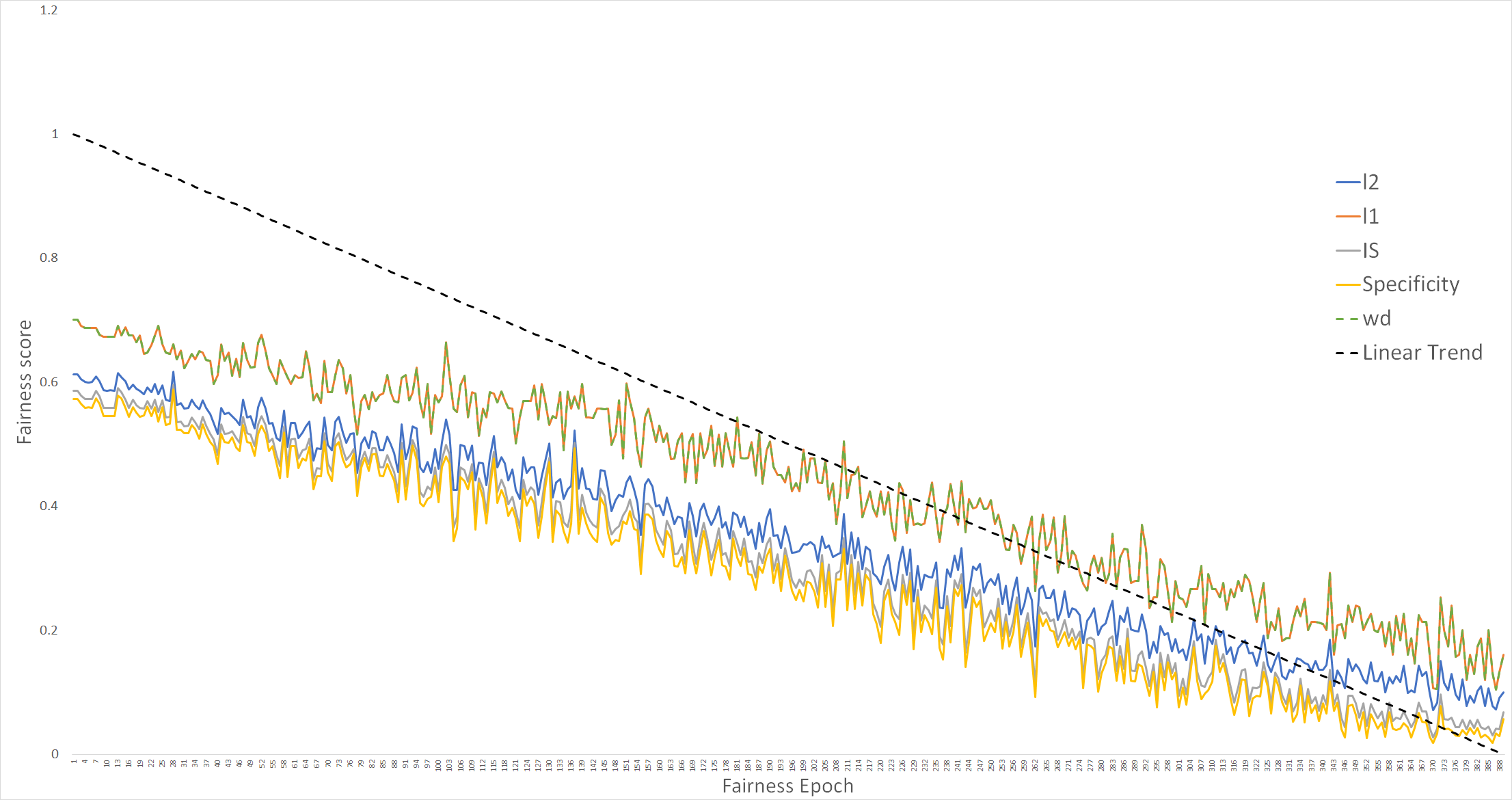}
    \caption{Normalise Fairness score with sweeping distribution from AB-EP to Fair-EP at k=16}
    \label{Study_Tree}
\end{figure*}

\begin{figure*}[!ht]
\centering
    \includegraphics[scale=0.2]{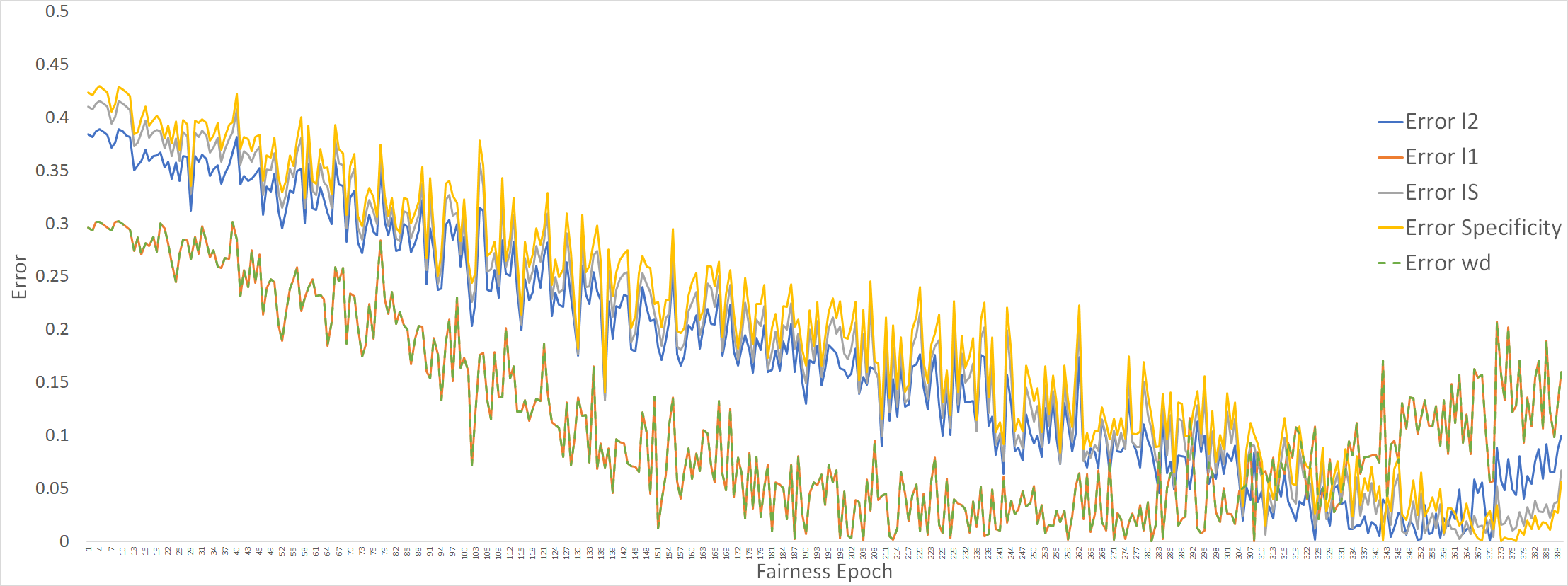}
    \caption{Error with sweeping distribution from AB-EP to Fair-EP at k=16}
    \label{Study_Tree}
\end{figure*}

\FloatBarrier
\section{Internal variability}

We observe that as mentioned in the trend line sweep that the metric generally trend downwards with increase fairness in the distribution. We ran the trend-line sweep starting from the respective AB-EP, e.g. \{1,0,0,0\} , \{0,1,0,0\}, \{0,0,1,0\} and \{0,0,0,1\} and measured the standard deviation. As per Figures 21,27, we observe that as the fairness epoch increase there is a decrease in the variability of the metric which we deem the internal-variability effect. This observation is in addition to the general decrease in the fairness score on all metrics. Fig 22,23,24,25 shows the individual metrics and their normalised scores at each fairness epoch. \\ \\
Fig 28 further demonstrates the internal-variability effect by showing that the error converges when at a lower point when comparing the AB-EP with the highest accuracy $EP_5\; Accuracy=$0.65  against $EP_7\; Accuracy=$0.87. Furthermore, when looking at the l1 metric on the two extreme accuracies in Fig 28 and 29, we see the different impacts that the internal variability effect has on the various extreme points. $EP_5$ in orange having the worst accuracy observe a steep improvement in its error. Whereas $EP_7$ having the highest accuracy observe a slower gradual improvement.

\begin{figure*}[!ht]
\centering
     \includegraphics[scale=0.2]{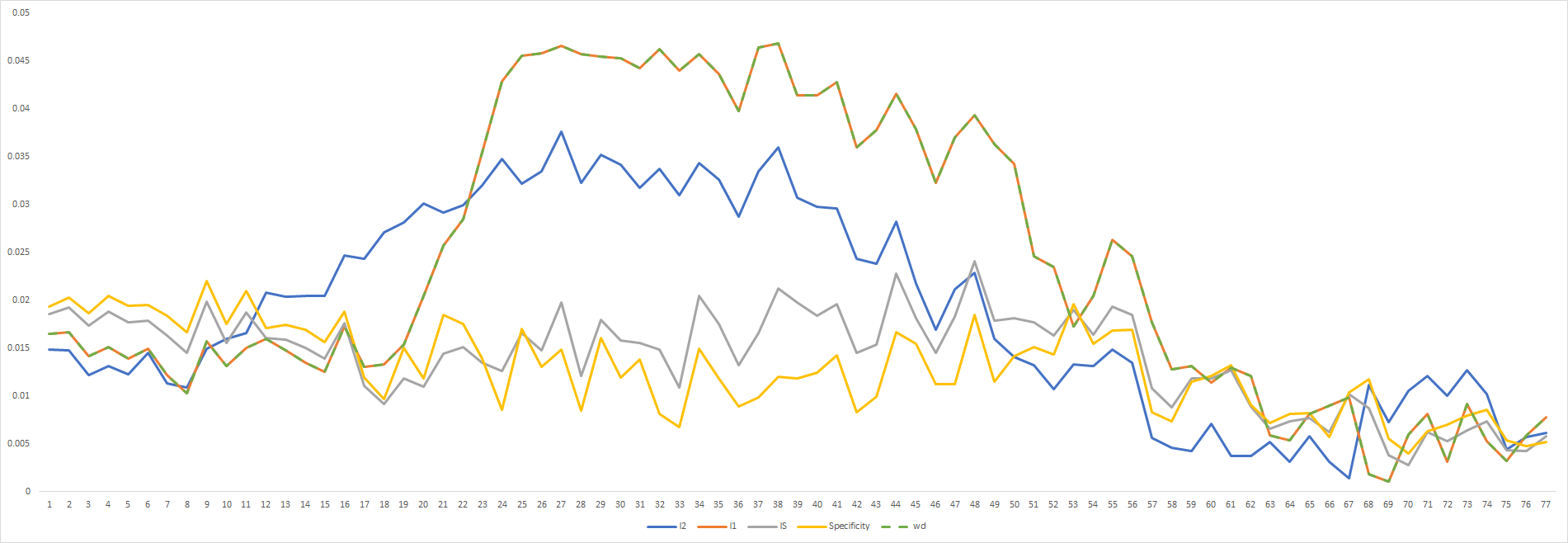}
    \caption{$k=4$, standard deviation starting from 4 different AB-EP to uniform distribution }
    \label{Study_Tree}
\end{figure*}

\begin{figure*}[!ht]
\centering
    \includegraphics[scale=0.2]{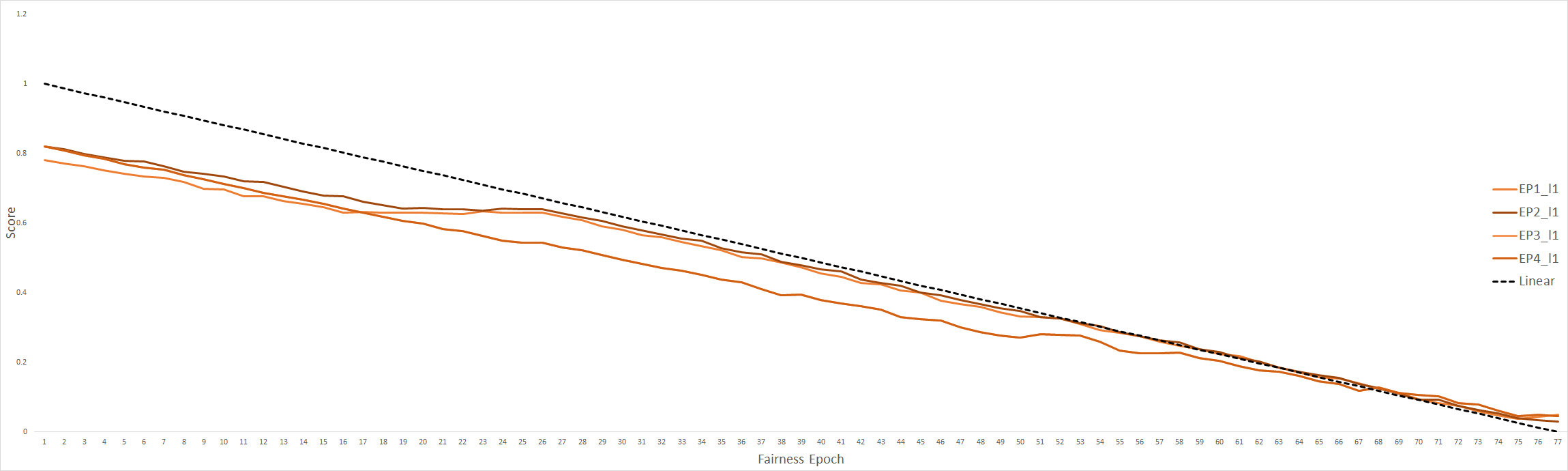}
    \caption{$k=4$, 4 extreme points trend line, L1 metric }
    \label{Study_Tree}
\end{figure*}

\begin{figure*}[!ht]
\centering
    \includegraphics[scale=0.2]{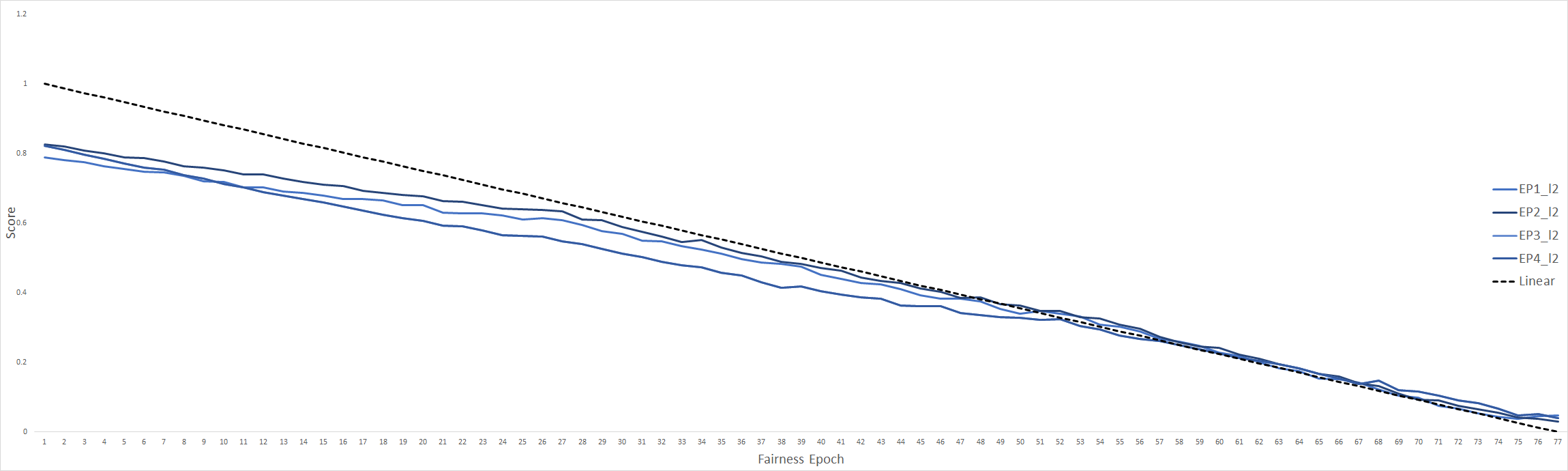}
    \caption{$k=4$, 4 extreme points trend line, L2 metric }
    \label{Study_Tree}
\end{figure*}

\begin{figure*}[!ht]
\centering
    \includegraphics[scale=0.2]{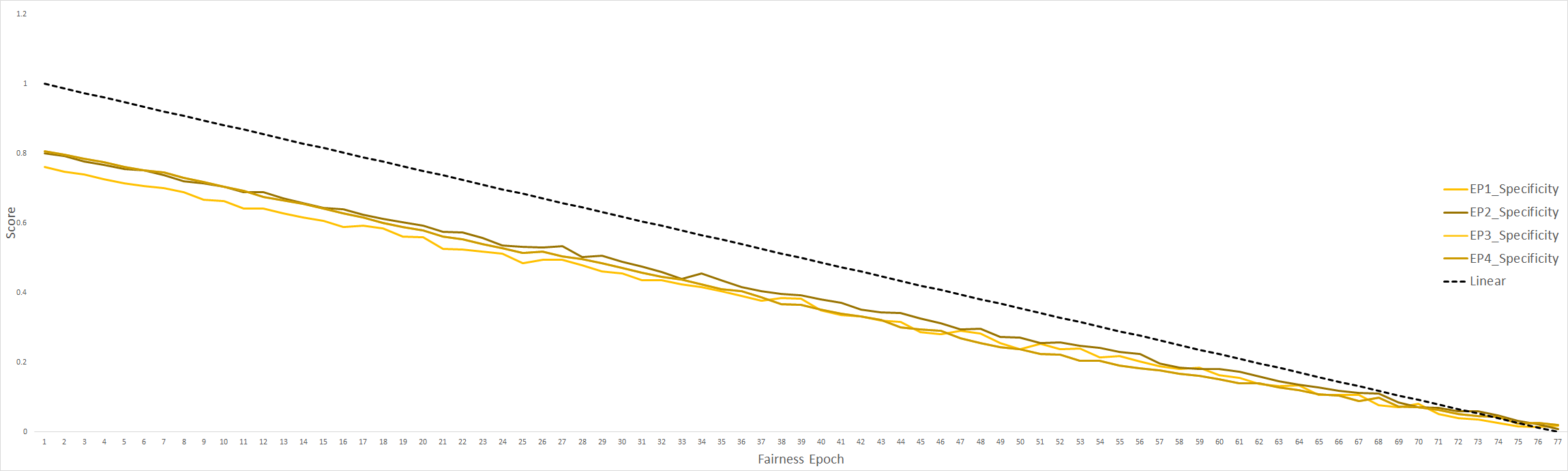}
    \caption{$k=4$, 4 extreme points trend line, Specificity metric }
    \label{Study_Tree}
\end{figure*}

\begin{figure*}[!ht]
\centering
    \includegraphics[scale=0.2]{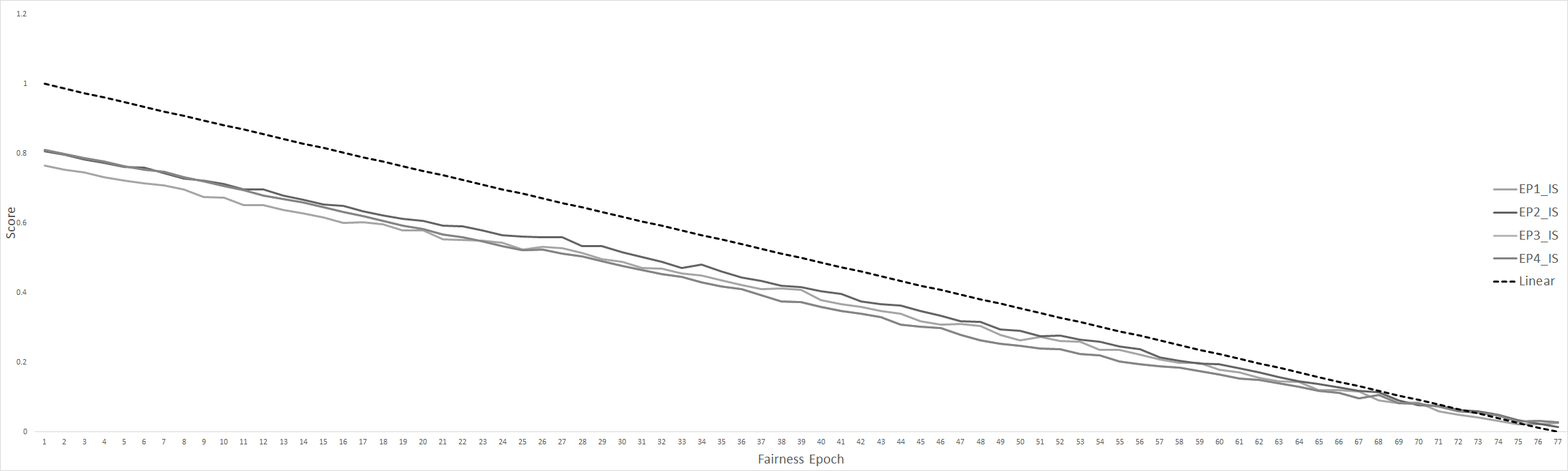}
    \caption{$k=4$, 4 extreme points trend line, IS metric }
    \label{Study_Tree}
\end{figure*}

\begin{figure*}[!ht]
\centering
    \includegraphics[scale=0.3]{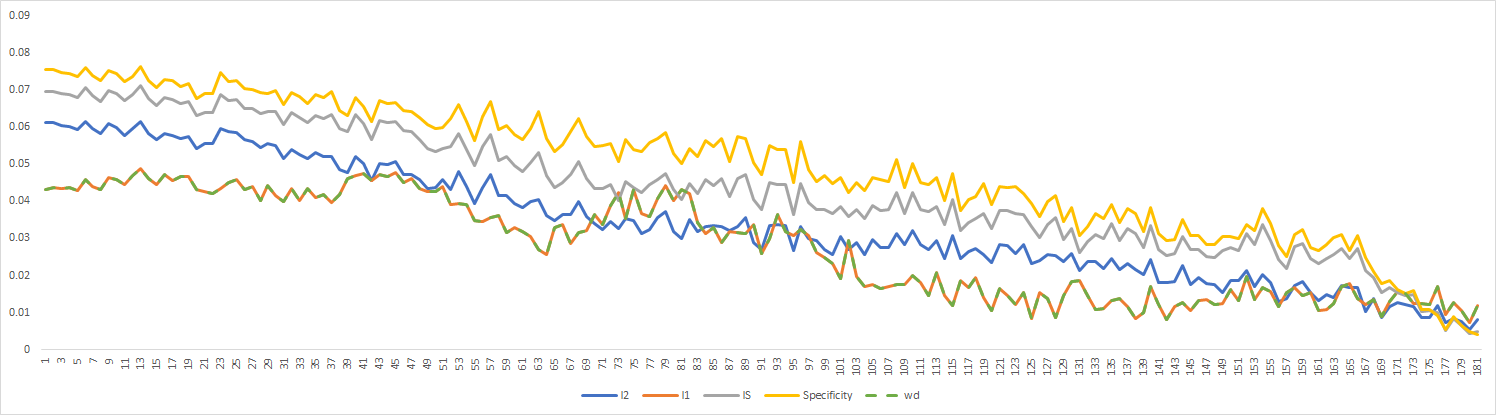}
    \caption{$k=8$, standard deviation starting from 8 different AB-EP to uniform distribution }
    \label{Study_Tree}
\end{figure*}

\begin{figure*}[!ht]
\centering
    \includegraphics[scale=0.25]{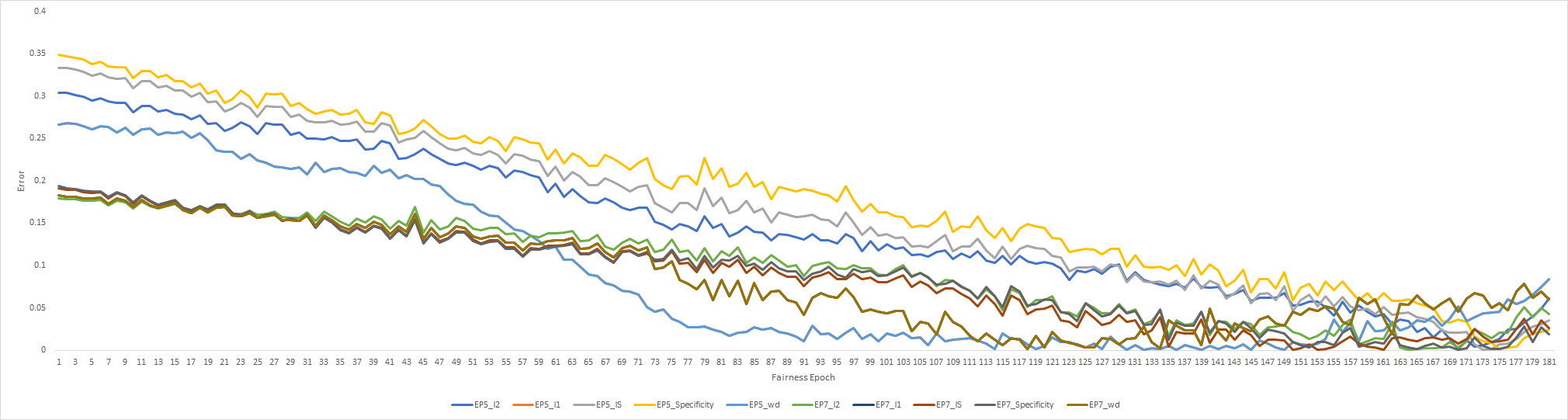}
    \caption{$k=8$, Max and Min accuracy attributes sweep from AB-EP to fair-EP }
    \label{Study_Tree}
\end{figure*}

 %L1 Max and Min
\begin{figure*}[!ht]
\centering
    \includegraphics[scale=0.25]{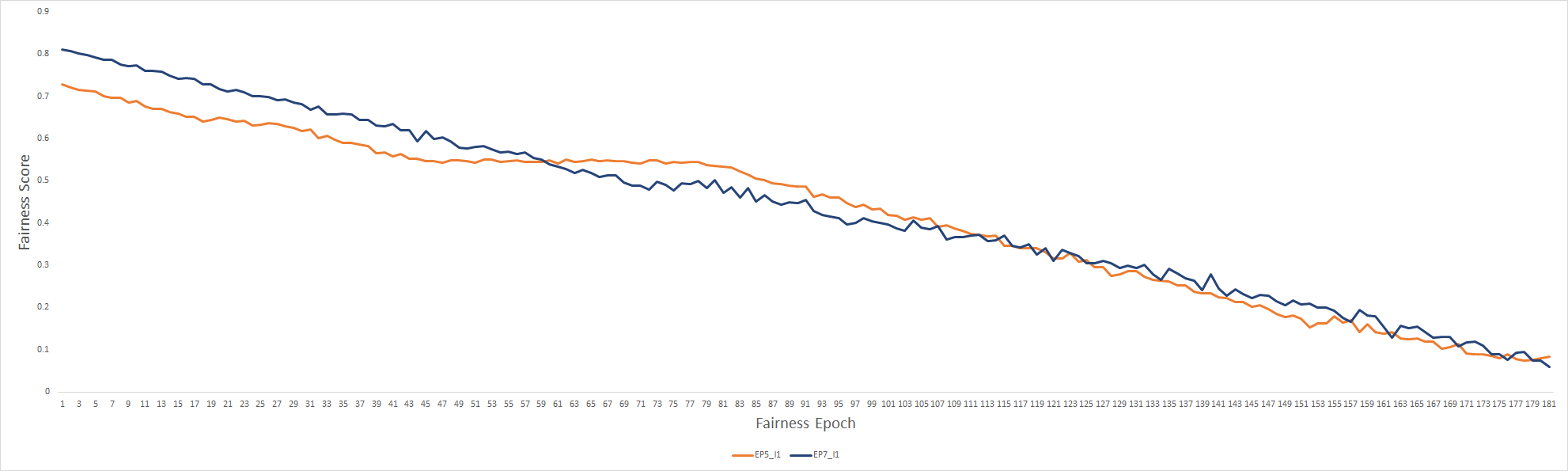}
    \caption{$k=8$, Max and Min accuracy attributes sweep from AB-EP to fair-EP for L1}
    \label{fig:maxminsweep8l1score}
\end{figure*}

\begin{figure*}[!ht]
\centering
    \includegraphics[scale=0.25]{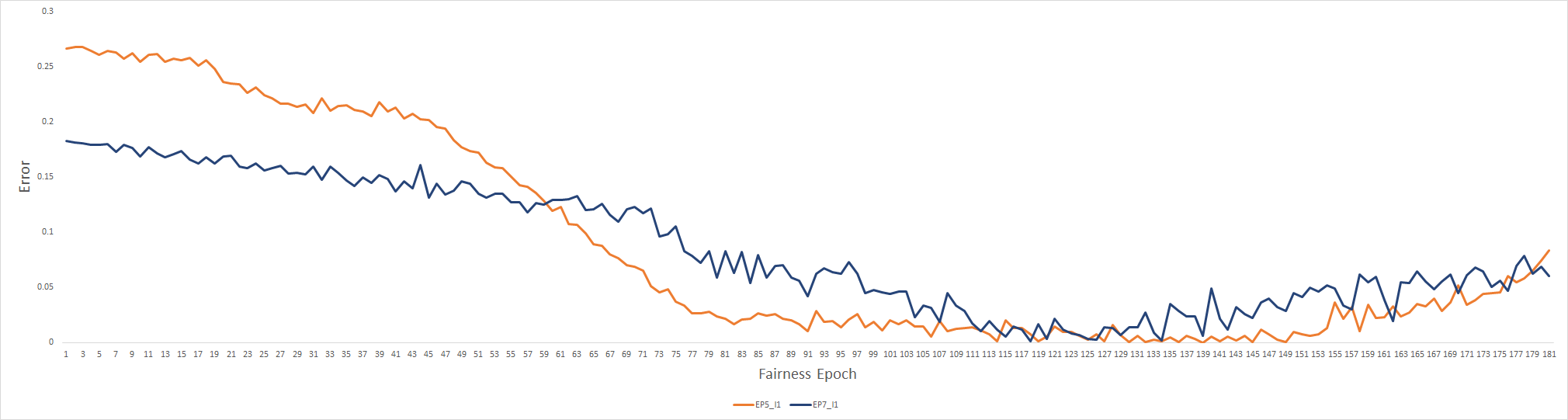}
    \caption{$k=8$, Max and Min accuracy attributes sweep from AB-EP to fair-EP for L1 }
    \label{fig:maxminsweep8l1error}
\end{figure*}

\FloatBarrier
%\newpage
%\section{Sampling method}
%Given the $Attribute=\{Male, Black\,hair\}$ we first pre-process the training data into the following categories $\bu=\{[0,0],[0,1],[1,0],[1,1]\}$. Where $[0,0]$ represents the permutation of female without black hair. We then attain the respective attribute counts available for each $A_{ti}$. For this example [0. 0.] count=9372,  [0. 1.]  count=2875 ,[1. 0.] count=5168 , [1. 1.] count=2547 in the celebA data set. \\ \\
%For proper sampling to occur, we need to ensure that both the extreme points, AB-EP and Fair-EP, can be sampled. In addition, the number of samples should not change per experiment. Any other distribution would fall between the range of the two EP. we reduce the conditions required from the notion that if all AB-EP can be sampled then the Fair-EP can be sampled. Lastly, for a consistent experiment, we need to ensure that all AB-EPs have the same sample count. Hence, we require $Min(AB-EP)=Min(\bu \, count)=2547$. \\ \\
%We then randomly sample 2547 from each attribute in $\bu$ to form the new $D_{bias}$ . During the experiments distribution are then randomly sampled from the new $D_{bias}$. For example, sampled distribution $P_{generated}=\{0.8,0.1,0.05,0.05\}$ we randomly sample $D_{generated\_Count}=\{2038,255,127,127\}$ from our $D_{bias}$ .

%\begin{figure}[!ht]
%\centering
%    \includegraphics[scale=0.25]{Figures/Sampling methods/Sampling_method.png}
%    \caption{Sampling Illustration}
%    \label{Study_Tree}
%\end{figure}

\end{appendices}

\end{document}